\pdfoutput=1

\documentclass[11pt]{article}

\usepackage[final]{EMNLP2023}
\usepackage{times}
\usepackage{latexsym}

\usepackage[T1]{fontenc}

\usepackage[utf8]{inputenc}

\usepackage{microtype}

\usepackage{inconsolata}

\usepackage{graphicx} 
\usepackage{enumitem}
\usepackage{multirow}
\usepackage{booktabs}
\usepackage{url}
\usepackage{amsfonts}
\usepackage{amsmath}
\usepackage{graphicx} 

\usepackage{algpseudocode}
\usepackage[vlined, ruled,linesnumbered]{algorithm2e}

\title{Instructions for EMNLP 2023 Proceedings}

\title{Crafting Personalized Agents through Retrieval-Augmented Generation on Editable Memory Graphs}

\author{Zheng Wang\textsuperscript{1}, Zhongyang Li\textsuperscript{1}, Zeren Jiang\textsuperscript{1}, Dandan Tu\textsuperscript{1}, Wei Shi\textsuperscript{1}\\
  \textsuperscript{1}Huawei Technologies, Co., Ltd.\\
\texttt{\{wangzheng155,lizhongyang6,jiangzeren2,tudandan,w.shi\}@huawei.com}}

\begin{document}
\maketitle
\begin{abstract}

In the age of mobile internet, user data, often referred to as memories, is continuously generated on personal devices. Effectively managing and utilizing this data to deliver services to users is a compelling research topic. In this paper, we introduce a novel task of crafting personalized agents powered by large language models (LLMs), which utilize a user's smartphone memories to enhance downstream applications with advanced LLM capabilities. To achieve this goal, we introduce \texttt{EMG-RAG}, a solution that combines Retrieval-Augmented Generation (RAG) techniques with an Editable Memory Graph (EMG). This approach is further optimized using Reinforcement Learning to address three distinct challenges: data collection, editability, and selectability. Extensive experiments on a real-world dataset validate the effectiveness of \texttt{EMG-RAG}, achieving an improvement of approximately 10\% over the best existing approach. Additionally, the personalized agents have been transferred into a real smartphone AI assistant, which leads to enhanced usability.

\end{abstract}

\section{Introduction}
\label{sec:introduction}

In the era of mobile internet, personal information is constantly being generated on smartphones. This data, referred to as personal memories, is often scattered across everyday conversations with AI assistants (e.g., Apple's Siri), or within a user's apps (e.g., screenshots), including emails, calendars, location histories, travel activities, and more. As a result, managing and utilizing these personal memories to provide services for users becomes a challenging yet attractive task. With the emergence of advanced large language models (LLMs), new opportunities arise to leverage their semantic understanding and reasoning capabilities to develop personal LLM-driven AI assistants.

Motivated by this trend, we study the problem of crafting personalized agents that enhance the AI assistants with the capabilities of LLMs by leveraging users' memories on smartphones. Unlike existing personal LLM agents~\cite{li2024personal}, such as those designed for psychological counseling~\cite{zhong2024memorybank}, housekeeping~\cite{han2024llm}, and medical assistance~\cite{zhang2023memory}, the personalized agents face unique challenges due to practical scenarios and remains relatively unexplored in current methods. 

These challenges can be summarized below. 
\underline{(1) Data Collection}: Personal memories should encompass valuable information about a user. Extracting these memories from everyday trivial conversations presents unique challenges in data collection, especially considering that existing datasets like personalized chats sourced through crowdsourcing~\cite{zhang2018personalizing} or psychological dialogues~\cite{zhong2024memorybank} lack this property. Moreover, constructing annotated data, such as QA pairs, is essential for enabling effective training of personalized agents. 
\underline{(2) Editability}: Personal memories are dynamic and continuously evolving, requiring three types of editable operations: insertion, deletion, and replacement. For example, 1) insertion occurs when new memories are added; 2) deletion is necessary for time-sensitive memories, such as a hotel voucher that expires and needs to be removed; 3) replacement is required when an existing memory, such as a flight booking, undergoes a change in departure time and needs updating. Therefore, a carefully designed memory data structure is essential to support this editability. 
\underline{(3) Selectability}: To enable the memory data services for real-world applications, it often requires querying a combination of multiple memories. For example, in a QA scenario (illustrated in Table~\ref{tab:data_collection}), the AI assistant answering a question about ``a secretary's boss's flight departure time'' needs several memories: the secretary booked a flight to Amsterdam for her boss ($M_1$); the flight's number is EK349 ($M_2$); the departure time for EK349 is at 01:40 on 2024-05-12 ($M_4$). To achieve this, one intuitive approach is to use Retrieval-Augmented Generation (RAG)~\cite{lewis2020retrieval} to find relevant memories and form a context that is fed into a LLM to generate answers. Here, we discuss two potential solutions and their limitations, which motivate the proposed solution.
1) Needles in a Haystack (NiaH)~\cite{briakou2023searching}: it organizes all memories into a single context (the ``Haystack'') and inputs this into a LLM, relying on the capability of a LLM itself to identify relevant memories (the ``Needles'') for generating an answer. However, this method incurs significant overhead by extending the LLM's context window and introduces noise from irrelevant memories, hindering the LLM's ability to generate accurate answers.
2) Advanced RAG~\cite{wang2024mrag,DBLP:journals/corr/abs-2305-14283}: many advanced RAG techniques still rely on Top-$K$ retrieval to identify relevant memories. However, a fixed parameter $K$ may limit the LLM's ability to uncover all relevant memories, especially for the questions requiring diverse memory combinations. Thus, an adaptive selection mechanism is essential for the personalized applications.

To this end, we introduce a new solution called \texttt{EMG-RAG}, which presents the first attempt of its kind to address these challenges. We discuss the solution along with the rationales behind it below. 
\underline{For (1)}, we utilize a business dataset collected from a real AI assistant, which includes daily conversations with the assistant, and users' app screenshots, to extract personal memories. Specifically, we leverage the capabilities of GPT-4~\cite{OpenAI2023GPT-4} to clean the raw data into memories. We organize the memories chronologically, and then use GPT-4 to generate QA pairs within each session (a set of consecutive memories). We also tag the memories involved in generating these QA pairs, which are then used for subsequent training purposes.
\underline{For (2)}, we introduce a three-layer data structure, called Editable Memory Graph (EMG). The first two layers form a tree structure in accordance with the business scopes, while the third layer consists of a user's memory graph parsed from the memory data. This design is motivated by three considerations: 1) the tree structure allows for partitioned management of various memory categories, facilitating expansion to other categories; and 2) memory data is partitioned under different categories, with the graph structure to capture their complex relationships, and 3) this enables efficient retrieval to locate specific memories for editing, by searching within relevant partitions rather than the entire dataset.
\underline{For (3)}, we introduce a reinforcement learning (RL) agent that adaptively selects memories on the EMG, without being constrained to a fixed Top-$K$ approach. The rationale of using RL resembles a boosting process. Specifically, when the agent selects relevant memories (actions), it prompts a LLM (frozen) to generate improved answers. The quality of these answers is evaluated by a downstream task metric (reward), which then guides the agent to refine its policy for better memory selection. This results in an end-to-end optimization process aimed at achieving the desired goal for downstream tasks.

Overall, we make the following contributions. 
(1) We introduce a novel task of crafting LLM-driven personalized agents, leveraging users' personal memories to enhance their experience through LLM capabilities. This task differs from existing personal LLM agents in three key challenges: data collection, editability, and selectability.
(2) We propose \texttt{EMG-RAG}, a novel solution that combines EMG and RAG to address the three challenges. We show that it enables an end-to-end optimization process through reinforcement learning to achieve the goal of personalized agents.
(3) We conduct extensive experiments on a real-world business dataset across various LLM architectures and RAG methods for three downstream applications: question answering, autofill forms, and user services. Our approach demonstrates improvements of approximately 10.6\%, 9.5\%, and 9.7\% over the best existing approach for these tasks, respectively. Moreover, the personalized agents have been transferred into an AI assistant product, resulting in a notable improvement in user experience.
\section{Related Work}
\label{sec:related}

\textbf{Personalized Dialogue System.} To develop a personalized dialogue system (PDS), the PersonaChat dataset~\cite{zhang2018personalizing} is collected through crowdsourcing, which comprises Personas (each persona is defined by a set of profile sentences) and Chats (each chat is collected by two crowdworkers with two randomly assigned personas). Based on the dataset, various techniques have been studied to address challenges in PDS, including mutual persona perception~\cite{liu2020you,xu2022cosplay,kim2020will}, persona-sparsity~\cite{song2021bob,welch2022leveraging}, long-term persona memory~\cite{xu2022long,zhong2024memorybank}, etc. 
For example, $\mathcal{P}^2$BOT~\cite{liu2020you} is a GPT-based framework~\cite{radford2018improving}, specifically designed to enrich personalized dialogue generation through mutual persona perception. It aims to model the underlying understanding, such as character traits, within a conversation to facilitate mutual acquaintance between interlocutors. 
In addition, a PDS can be further enhanced by integrating internal reasoning techniques~\cite{hongru2023cue} or external acting techniques~\cite{wang2023large}, which aim to generate more personalized and factual responses. In this study, we construct user-personalized agents using practical memory data gathered from smartphone AI assistants. Leveraging these agents, we introduce three distinct applications: question answering, autofill forms, and user services.

\smallskip
\noindent\textbf{Retrieval-Augmented Generation on Knowledge Graph.} We review the literature on RAG on knowledge graphs across various tasks, including KBQA~\cite{ye2021rng,das2021case,wang2023keqing,shu2022tiara}, open-domain scenarios~\cite{yang2023enhancing}, table-related tasks~\cite{jiang2023structgpt}, human-machine conversation~\cite{zhang2020grounded}, and image captioning~\cite{hu2023reveal}. This paper~\cite{zhao2024retrieval} provides a detailed survey on these tasks with RAG techniques. 
Specifically, TIARA~\cite{shu2022tiara} stands out as a KBQA model employing multi-grained retrieval (entities, logical forms, and schema items) from knowledge graphs. This approach aids pre-trained language models in mitigating generation errors. 
In this study, we introduce a novel EMG structure to manage users' personal memories. Further, we employ RL to model the RAG process, which optimizes the memory selection on the graph.

\smallskip
\noindent\textbf{Model Editing.} Model editing represents a recent research area focused on correcting model predictions in light of evolving real-world dynamics. It edits the behavior of pre-trained language models within specific domains, and preserving performance across other domains without compromise.
Some existing methods~\cite{de2021editing,mitchell2021fast} employ learnable model editors, which are trained to predict the weights of the base model undergoing editing. Other methods~\cite{meng2022locating,meng2022mass,li2024pmet} are designed to identify stored facts (such as specific neurons in the network) and adjust corresponding activations to reflect changed facts. Additionally, SERAC~\cite{mitchell2022memory} utilizes an external memory to store edits, adaptively altering the base model's predictions by retrieving relevant edits. In our study, we leverage a LLM to focus on user personal memories rather than global knowledge. Additionally, we support dynamic user edits on the EMG and utilize RAG with a frozen LLM to respond to these changes.
\section{Problem Statement}
\label{sec:problem}

We study the problem of developing personalized agents for users on smartphone AI assistant platforms (such as Apple's Siri or Samsung's Bixby). These agents are designed to assist users in performing personalized tasks, requiring the fulfillment of the following two properties in practical scenarios:

\begin{itemize}[leftmargin=3mm]
    \item[-] Editability: The responses from the agents may be editable based on the users' dynamic memory data, which involves insertion, deletion, and replacement operations corresponding to different usage scenarios, as illustrated in Figure~\ref{fig:overall}(a).
    \item[-] Selectability: The agents can select relevant memories to respond to users' queries, with some queries requiring the combination of multiple memories to generate responses through a base language model, as illustrated in Figure~\ref{fig:overall}(b).
\end{itemize}

By satisfying these properties, the agents aim to enhance the user experience during interactions with their smartphone AI assistants. These agents offer essential functionalities to support personalized applications, including question answering, autofill forms, and user services like reminders for important events and times, and travel navigation (further details will be discussed in Section~\ref{sec:apps}). 

\begin{figure*}
  \small
  \centering
  \includegraphics[width=0.85\linewidth]{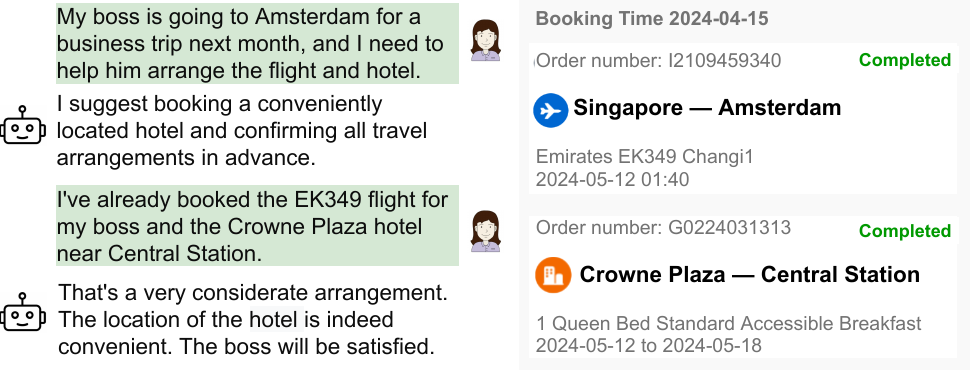}
  \vspace*{-3mm}
  \caption{An example of data collection. Step-1: Raw data is gathered on smartphone AI assistant platforms, e.g., everyday conversations between users and assistants, and the extraction of app screenshot contents through OCR.}
  \label{fig:data_collection}
  \vspace*{-3mm}
\end{figure*}

\begin{table*}[]
\small
\centering
\setlength{\tabcolsep}{2pt}
\caption{An example of data collection. Step-2: GPT-4 generates memories from raw data. Step-3: GPT-4 forms QA pairs using several memories, and produces the required memories, which are utilized for training the \texttt{EMG-RAG}.}
\vspace*{-3mm}
\label{tab:data_collection}
\begin{tabular}{l|l}
\hline
\multicolumn{1}{c|}{Step-2: Memories (generated by GPT-4)}                                                                                       & \multicolumn{1}{c}{Step-3: QA pairs with memories (generated by GPT-4)}                                                                                                                                                        \\ \hline
\begin{tabular}[c]{@{}l@{}}$M_1$: My boss is traveling to Amsterdam next month, \\ I assist with flight and hotel arrangements.   \end{tabular} & \multirow{3}{*}{\begin{tabular}[c]{@{}l@{}}$Q$: What time is my boss's flight to Amsterdam? \\ $A$: Your boss flight EK349 departs at 01:40 on 2024-05-12. \\ Required memories: $M_1, M_2, M_4$\end{tabular}}                                    \\

$M_2$: I booked the EK349 flight.                                                                                                            &                                                                                                                                                                                                                        \\
$M_3$: I booked the Crowne Plaza near Central Station.                                                                                       &                                                                                                                                                                                                                        \\ \cline{2-2} 
$M_4$: The EK349 flight departs at 01:40 on 2024-05-12.                                                                                       & \multirow{3}{*}{\begin{tabular}[c]{@{}l@{}}\\$Q$: When dose the hotel I booked for my boss start and end? \\ $A$: The Crowne Plaza reservation is from 2024-05-12 to 2024-05-18. \\ Required memories: $M_1, M_3, M_5$\end{tabular}} \\
\begin{tabular}[c]{@{}l@{}}$M_5$: The Crowne Plaza reservation is for \\ 2024-05-12 to 2024-05-18.  \end{tabular}                              &                                                                                                                                                                                                                        \\
\begin{tabular}[c]{@{}l@{}}$M_6$: The Crowne Plaza reservation includes a Queen \\ Bed Standard Accessible room with breakfast. \end{tabular} &                                                                                                                                                                                                                        \\ \hline
\end{tabular}
\vspace*{-4mm}
\end{table*}

\section{Methodology}
\label{sec:method}

\subsection{Data Collection}
\label{sec:dc}

The process entails (1) gathering raw data, such as everyday conversations or screenshots from user interactions with the smartphone AI assistants; (2) extracting crucial information from this raw data, referred to as memories (denoted by $M$); and (3) generating QA pairs (denoted by $<Q,A>$), and outputting the required memories to facilitate this pairing. For (1), we acquire data from real AI assistant products and employ text processing techniques like OCR to extract content from screenshots. Subsequently, for (2) and (3), we leverage the capabilities of LLMs, such as GPT-4~\cite{OpenAI2023GPT-4}, to extract key memories from the raw data and create QA pairs. These pairs serve the purpose of training personalized agents for the proposed \texttt{EMG-RAG}. To illustrate the collection process, we provide a running example in Figure~\ref{fig:data_collection} and Table~\ref{tab:data_collection}, which involve the three primary steps. Further details are outlined in Appendix~\ref{asec:data_collection}.

We discuss the rationales of the data collection. First, as a user's personalized agent integrated within the smartphone AI assistant, the conversations and screenshots provide natural data sources for crafting these agents. Second, leveraging GPT-4's language generation capabilities enables us to generate a wide range of memories from the raw data, significantly reducing manual effort. Third, the involved memories and collected QA pairs serve as labels to supervise the training of the retrieval and generation processes in our framework.

\begin{figure*}
  \centering
  \includegraphics[width=0.99\linewidth]{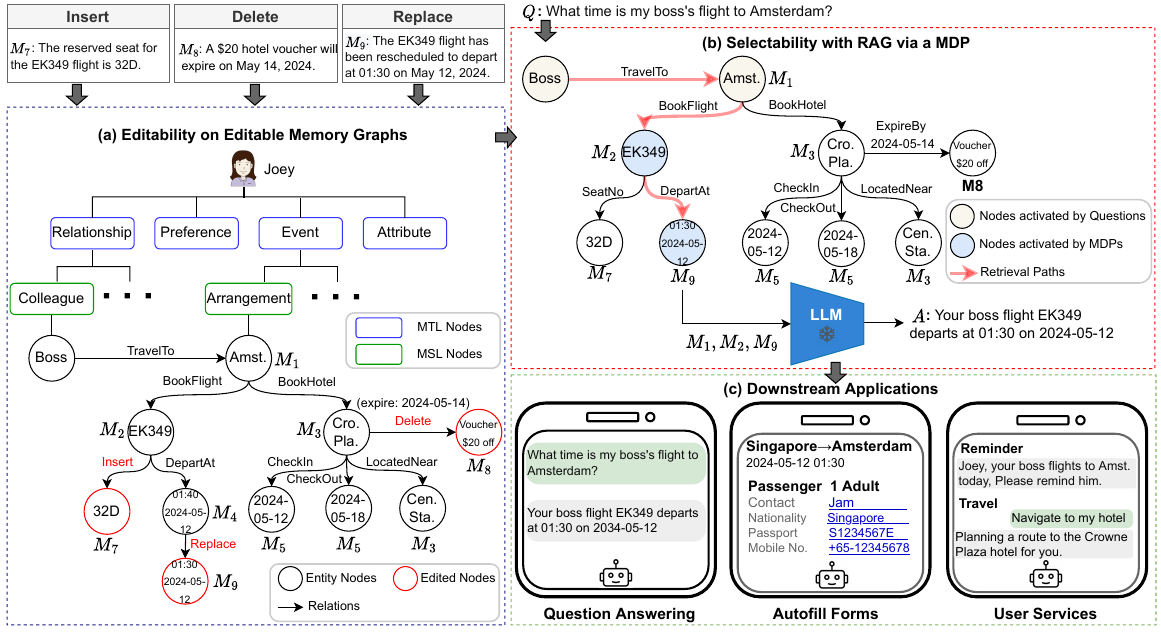}
  \vspace*{-2mm}
  \caption{The architecture of the proposed \texttt{EMG-RAG}, demonstrated with the running example in data collection (Section~\ref{sec:dc}). It supports three editability operations: insertion (e.g., $M_7$), deletion (e.g., $M_8$), and replacement (e.g., $M_9$), based on the EMG structure (Section~\ref{sec:emg}). Subsequently, the edited EMG undergoes RAG to select relevant memories (e.g., $M_1,M_2,M_9$) for a given question $Q$ via a MDP (Section~\ref{sec:mdp}). The generated answers $A$ by a frozen LLM further facilitates three downstream applications (Section~\ref{sec:apps}).}\label{fig:overall}
  \vspace*{-4mm}
\end{figure*}

\subsection{Editable Memory Graphs}
\label{sec:emg}

\textbf{The EMG Construction and Insights.} Utilizing a user's memories, we establish the Editable Memory Graph with a multilayered structure, depicted in Figure~\ref{fig:overall}(a), where the user is the root node.

\underline{Memory Type Layer (MTL)}: Aligned with the business scope, we categorize memories into 4 predefined types: Relationship, Preference, Event, and Attribute. Details are provided in Appendix~\ref{asec:MSL}.

\underline{Memory Subclass Layer (MSL)}: The MSL further outlines subclasses for each type, where the MTL and MSL are organized in a hierarchical tree structure to manage the memories. Detailed subclasses with examples are listed in Appendix~\ref{asec:MSL}.

\underline{Memory Graph Layer (MGL)}: The memory graph is built by utilizing the collected memories, employing entity recognition for nodes and relation extraction for edges. 
In this graph, each in-degree node is associated with its corresponding memory, e.g., the in-degree node (01:40 on 2024-05-12) contains $M_4$, as shown in Figure~\ref{fig:overall}(a). 
Further, to establish the connection between the MSL and MGL, TransE embeddings~\cite{bordes2013translating} are employed to capture semantic information of nodes in MSL (subclasses) and MGL (entities), respectively. Then, each entity is assigned to its closest classes based on these embeddings. It is noteworthy that entity nodes are categorized into different subclasses, and their connections may span across different classes, e.g., ``Boss'' and ``Amsterdam'' are linked across ``Colleague'' and ``Arrangement'' classes in Figure~\ref{fig:overall}(a). This design enables further traversal across various parts of the whole graph.

We discuss the insights of the EMG construction: 1) the tree hierarchy (MTL and MSL) offers a partitioned memory management approach, to facilitate the expansion of additional types and subclasses in accordance with business needs; 2) the entity nodes and corresponding memories are organized into separate subclass partitions, with the graph structure (MGL) to capture their complex relationships between memories; 3) it enables efficient retrieval of memories for further editing operations by first locating a relevant partition, e.g., querying partition centers (the mean of the memory embeddings), instead of searching through all memories.

\smallskip
\noindent\textbf{The EMG Editing.} When editing a given memory within the EMG (e.g., insertion, deletion, or replacement), the process involves three steps. Initially, a model such as CPT-Text~\cite{neelakantan2022text} is employed to acquire memory representations. Then, the memory is assigned to its nearest subclass (partition), and the Top-1 retrieved memory within the partition is then returned, and editing operations are performed based on comparing the relations between the given memory and the retrieved memory. Specifically, as illustrated in Figure~\ref{fig:overall}, (1) Insertion: It introduces a new relation to be added, e.g., obtaining a new memory containing flight seat number. (2) Deletion: It introduces a new relation, but it is valid for a specific period of time. e.g., a hotel voucher will expire on May 14, 2024. (3) Replacement: It provides an existing relation, and updates the corresponding entity nodes based on this relation, e.g., changing the departure time to 01:30 on May 12, 2024.

\subsection{MDP for Selecting Memories on EMGs}
\label{sec:mdp} 
Next, we outline the task of selecting memories based on an edited EMG. To achieve this, we employ an agent to traverse the EMG. Specifically, given a question $Q$, the agent selects a set of memories from the EMG denoted by $\mathbb{M} = \{M_i\}$, where $1 \le i \le |\mathbb{M}|$. The question $Q$ and memory set $\mathbb{M}$ are concatenated to generate an answer $\hat{A} \leftarrow \text{LLM}(Q \oplus \mathbb{M})$ using a LLM. We assess the generation quality using $\Delta(\hat{A}, A)$, where $A$ represents the collected ground truth answer for $Q$, and $\Delta(\cdot, \cdot)$ denotes a specific metric (e.g., ROUGE~\cite{lin-2004-rouge} or BLEU~\cite{DBLP:conf/wmt/Post18}). We note that a high-quality answer $\hat{A}$ benefits from the selected memories $\mathbb{M}$, which can then provide feedback with $\Delta(\cdot, \cdot)$ for subsequent selections. As a result, it iterates in a boosting process, and we optimize it using reinforcement learning. The environment, states, actions, and rewards are introduced below.

\smallskip
\noindent\textbf{Constructing Environment (Nodes activated by Questions).} Given an EMG, which often contains numerous memories in practice. Here, we confine the movement of the RL agent to a subset of memories to facilitate more focused selection. To achieve this, we first retrieve Top-$K$ memories for a given question $Q$, and based on these memories, we activate the corresponding nodes on the EMG (e.g., the nodes highlighted in yellow in Figure~\ref{fig:overall}(b)). Subsequently, the agent's traversal starts from each activated node via depth-first search.

\smallskip
\noindent\textbf{Modeling Memory Selection (Nodes activated by MDPs).} We model the graph traversal process as a MDP, involving states, actions, and rewards. 

\underline{States}: In the context where we have an input question $Q$, and visit a node $N_{G}$ (associated with a memory $M_i$ to be included into $\mathbb{M}$), and its relation $R_{G}$ on the EMG. We first extract the entity $N_{Q}$ and relation $R_{Q}$ from the $Q$, and the state $\mathbf{s}$ is defined by three cosine similarities $C(\cdot,\cdot)$, i.e.,

\begin{align}
\label{eq:state}
\mathbf{s} = \{C(\mathbf{v}_{N_{Q}}, \mathbf{v}_{N_{G}}), C(\mathbf{v}_{R_{Q}},\mathbf{v}_{R_{G}}), C(\mathbf{v}_Q, \mathbf{v}_{M_i})\}, 
\end{align}
where $\mathbf{v}_{\cdot}$ denotes the embedding vector for entities, relations, questions, or memories.

\underline{Actions}: We denote an action as $a$, and it has two choices during the graph traversal: including the visiting memory $M_i$ into $\mathbb{M}$, and searching its connected nodes; or stopping the current search, and restarting a search from other branches. Thus, the action $a$ is defined as: 
\begin{align}
\label{eq:action}
a = 1 \text{ (including)}\ \text{or} \ 0 \text{ (stopping)}.
\end{align}    
Consider the consequence of performing an action, it transitions the environment to the next state $\mathbf{s}'$, and affects which memory to be selected for constructing the state.

\underline{Rewards}: We denote the reward as $r$, which corresponds to the transition from the current state $\mathbf{s}_t$ to the next state $\mathbf{s}_{t+1}$ after taking action $a_t$. Specifically, when a memory $M$ is selected into $\mathbb{M}$, the generated answer by a LLM changes from $\hat{A}$ to $\hat{A}'$ accordingly. The quality of the generated answer $\hat{A}$ is evaluated using a specific metric $\Delta(\cdot,\cdot)$ (e.g., ROUGE or BLEU), and the reward $r$ is defined as: 
\begin{align}
\label{eq:reward_r}
r = \Delta(\hat{A}', A)-\Delta(\hat{A}, A),
\end{align}
where $A$ denotes the ground truth answer. We note that the objective of the MDP, which aims to maximize cumulative rewards, aligns with the goal of discovering memories to answer the question. To illustrate, consider a process through a sequence of states: $\mathbf{s}_1, \mathbf{s}_2, ..., \mathbf{s}_N$, concluding at $\mathbf{s}_N$. The rewards received at these states, except for the termination state, can be denoted as $r_1, r_2, ..., r_{N-1}$. When future rewards are not discounted, we have:
\begin{equation}
    \begin{aligned}
    \label{eq:rwd}
    \sum_{t=2}^{N}r_{t-1} &=\sum_{t=2}^{N}(\Delta(\hat{A}_{t}, A) - \Delta(\hat{A}_{t-1}, A)) \\
    &= \Delta(\hat{A}_N, y) - \Delta(\hat{A}_1, y),
\end{aligned}
\end{equation}
where $\Delta(\hat{A}_N, y)$ corresponds to the result of the final answer found throughout the entire iteration, and $\Delta(\hat{A}_1, y)$ represents an initial result that remains constant. Therefore, maximizing cumulative rewards is equivalent to maximizing the quality of the final generated answer. 

\smallskip
\noindent\textbf{Training Policies of MDPs.} Training the MDP policy involves two stages: warm-start stage (WS) and policy gradient stage (PG). \underline{In WS}, we employ supervised fine-tuning to equip the agent with the basic ability to select memories given a question $Q$. Specifically, based on a state $\mathbf{s}$, the agent undergoes a binary classification task to predict whether the memory $M_i$ should be included. This prediction is supervised according to whether the memory falls into the required memories (presented in the Step-3 in Table~\ref{tab:data_collection}). Thus, the objective is trained with binary cross-entropy, formulated as:
\begin{align}
\label{eq:mlLoss}
\mathcal{L}_\text{WS} = - y * \log(P) + (y - 1) * \log(1 - P),
\end{align}
where $y$ denotes the label (1 if the memory falls into the required memory set, and 0 otherwise), and $P$ is the predicted probability of the positive class.

\underline{In PG}, our main objective is to develop a policy $\pi_{\theta}(a|\mathbf{s})$ that guides the agent in selecting actions $a$ based on constructed states $\mathbf{s}$, aiming to maximize the cumulative reward $R_N$. We utilize the REINFORCE algorithm~\cite{williams1992simple, silver2014deterministic} for learning this policy, where the neural network parameters are denoted by $\theta$. The loss function is formulated as:

\begin{equation}
\label{eq:policy}
\mathcal{L}_\text{PG} = -R_N\ln\pi_{\theta}(a|\mathbf{s}).
  \end{equation}

\smallskip
\noindent\textbf{Inference Stage of \texttt{EMG-RAG}.} As shown in Figure~\ref{fig:overall}, the inference involves three steps: (a) collecting newly recorded memories from users and editing their EMGs; (b) using the edited EMGs to traverse the graph and retrieve relevant memories for LLM generation; (c) integrating the generated answers to serve users across three downstream applications.

\subsection{Discussion on Applications and Cold-start}
\label{sec:apps}

\smallskip
\noindent\textbf{Applications of the Personalized Agents.} As shown in Figure~\ref{fig:overall}(c), we explore the capabilities of personalized agents in three scenarios: (1) question answering, (2) autofill forms, and (3) user services. For (1), \texttt{EMG-RAG} can generate answers to users' questions when they interact with the smartphone AI assistants. For (2), the goal is to extract personal information from users' EMGs to automatically fill out various online forms, such as flight and hotel bookings.
To achieve this, we input form-related questions (e.g., ``What is the user's mobile number?'') into the LLM and use the generated entities to complete the forms. For (3), we focus on two specific domains. a) reminder service: It involves reminding users of recent events and times. To achieve this, we query a LLM for information about a user’s recent events and their associated times. b) travel service: We assist users with navigation by providing the address of a destination they might want to visit. Further, we integrate the generated answers (e.g., events, times, addresses) with external tools such as calendar or map apps to provide the services for users.

\smallskip
\noindent\textbf{Handling the Cold-start Problem.}
Given that \texttt{EMG-RAG} relies on generated questions for training, it may encounter a potential cold-start issue when deploying to answer real user questions. To address this issue, we utilize online learning to continuously fine-tune the agent using newly recorded questions and manually written answers, as outlined in Equation~\ref{eq:policy}. This approach aims to ensure that the model's policy remains up-to-date for online usage. We validate this method through online A/B testing, and the results demonstrate improvements in user experience, highlighting the positive impact of this strategy in practice.

\section{Experiments}

\subsection{Experimental Setup}
\label{sec:setup}

\noindent \textbf{Dataset and Ground Truth.} We conduct experiments on a real-world business dataset containing approximately 11.35 billion raw text data (including conversations and screenshot contents) from an AI assistant product collected between March 2024 and June 2024. After data cleaning, the dataset forms around 0.35 billion memories. We follow the data distribution to randomly sample 2,000 users for training and 500 users for testing.

As detailed in Section~\ref{sec:dc}, we establish the ground truth for the applications of question answering and autofill forms/user services using GPT-4 generated answers and key entities (e.g., identification number, address, and time), respectively. We provide a quality evaluation for the collected dataset in Section~\ref{sec:result}.




\noindent \textbf{Baselines.} We compare \texttt{EMG-RAG} with the following RAG methods. 1) NiaH~\cite{briakou2023searching}: It simply inputs all of the users' memories into a LLM within the context window size to generate the answer. 2) Naive~\cite{DBLP:journals/corr/abs-2305-14283}: It implements a basic RAG execution process involving indexing, retrieval, and generation. 3) M-RAG~\cite{wang2024mrag}: It partitions a database and employs Multi-Agent RL to train two agents for RAG. Agent-S selects a database partition, while Agent-R refines the stored memories within that partition to generate a better answer. In our adaptation, we omit Agent-R since, in our scenario, the generated answers must be grounded in the user's personal memories, which cannot be altered due to potential risks. 4) Keqing~\cite{wang2023keqing}: The knowledge graph-based method decomposes a question into sub-questions, retrieves candidate entities, generates answers for each sub-question, and then integrates them into a comprehensive final answer.

In addition, we integrate the RAG methods into three typical LLM architectures. 1) GPT-4~\cite{OpenAI2023GPT-4} is a Transformer-based pre-trained model known for its human-level performance. 2) ChatGLM3-6B~\cite{du2022glm} is a long-text dialogue model with a sequence length of 32K. 3) PanGu-38B~\cite{ren2023pangu} is a dialogue sub-model of the PanGu series, which follows a Mixture of Experts (MoE) architecture.

\noindent \textbf{Evaluation Metrics.} We evaluate the effectiveness of \texttt{EMG-RAG} in three downstream applications. For question answering, we assess the quality of generated answers with the ground truth, and reporting ROUGE (R-1/2/L)~\cite{lin-2004-rouge} and BLEU~\cite{DBLP:conf/wmt/Post18} scores. For autofill forms and user services, we generate key entities and report Exact Match (EM) accuracy. Overall, higher values (i.e., ROUGE, BLEU, EM) indicate better results~\footnote{We remark that all reported results are statistically significant, as confirmed by a t-test with $p<0.05$.}.


\begin{table*}[]
\caption{Effectiveness of \texttt{EMG-RAG} in downstream applications.}
\centering
\vspace{-3mm}
\label{tab:effectiveness}
\begin{tabular}{l|c|clcc|c|cc}
\hline
\multirow{2}{*}{LLM} & \multirow{2}{*}{RAG} & \multicolumn{4}{c|}{Question Answering} & \multirow{2}{*}{\begin{tabular}[c]{@{}c@{}}Autofill Forms\\ (EM)\end{tabular}} & \multicolumn{2}{c}{User Services (EM)} \\ \cline{3-6} \cline{8-9} 
                     &                      & R-1      & R-2     & R-L     & BLEU     &                                                                                & Reminder            & Travel           \\ \hline
GPT-4                & NiaH                 &79.89 &64.65 &70.66 &38.72 &84.86 &84.49 &94.81                 \\
GPT-4                & Naive                &70.87 &58.34 &66.82 &46.65 &78.40 &85.34 &94.52                 \\
GPT-4                & M-RAG                &88.71 &77.18 &84.74 &64.16 &90.87 &93.75 &86.67                  \\
GPT-4                & Keqing               &72.11 &57.19 &65.46 &35.89 &82.03 &90.17 &72.71                  \\ \hline
GPT-4                & \texttt{EMG-RAG}     &\textbf{93.46} &\textbf{83.55} &\textbf{88.06} &\textbf{75.99} &\textbf{92.86} &\textbf{96.43} &\textbf{91.46}                  \\
ChatGLM3-6B          & \texttt{EMG-RAG}     &85.31 &76.03 &82.32 &56.88 &85.71 &87.50 &81.25                  \\
PanGu-38B            & \texttt{EMG-RAG}     &91.64 &82.86 &86.71 &75.11 &90.99 &96.41 &89.05                 \\ \hline
\end{tabular}
\vspace{-3mm}
\end{table*}


\begin{table*}[]
\caption{Effectiveness of \texttt{EMG-RAG} for continuous edits.}
\centering
\setlength{\tabcolsep}{3pt}
\vspace{-3mm}
\label{tab:duration}
\begin{tabular}{l|ccc|ccc|ccc|ccc}
\hline
Duration (weeks) & \multicolumn{3}{c|}{1} & \multicolumn{3}{c|}{2} & \multicolumn{3}{c|}{3} & \multicolumn{3}{c}{4} \\ \hline
\# of edits          & \multicolumn{3}{c|}{2,515}  & \multicolumn{3}{c|}{9,644}  & \multicolumn{3}{c|}{2,096}  & \multicolumn{3}{c}{6,290}  \\ \hline
Apps (GPT-4)     & QA     & AF    & US    & QA     & AF    & US    & QA     & AF    & US    & QA    & AF    & US    \\ \hline
M-RAG            &88.48         &91.67       &90.28       &86.39        &88.89       &89.39       &85.31        &87.50       &87.83       &85.09    &83.33    &83.21     \\
\texttt{EMG-RAG} &\textbf{95.38}         &\textbf{93.75}        &\textbf{93.67}       &\textbf{96.93}       &\textbf{95.83}        &\textbf{95.89}       &\textbf{94.53}       &\textbf{96.88}        &\textbf{96.99}     &\textbf{94.99} &\textbf{97.50} &\textbf{97.54}        \\ \hline
\end{tabular}
\vspace{-5mm}
\end{table*}

\begin{table}[]
\caption{Ablation study.}
\centering
\setlength{\tabcolsep}{4.5pt}
\vspace{-3mm}
\begin{tabular}{l|cccc}
\hline
\multicolumn{1}{l|}{Components} & R-1 & R-2 & R-L & BLEU \\ \hline
\multicolumn{1}{l|}{\texttt{EMG-RAG}}    &\textbf{93.46} &\textbf{83.55} &\textbf{88.06} &\textbf{75.99}      \\ \hline
w/o Act. Nodes    &90.96     &82.72     &86.13     &65.07      \\
w/o WS            &92.95     &82.52     &86.49     &69.13      \\
w/o PG            &90.59     &80.69     &86.19     &65.65      \\ \hline
\end{tabular}
\vspace{-4mm}
\end{table}

\noindent \textbf{Implementation Details.} We implement \texttt{EMG-RAG} and other baselines in Python 3.7, using the Faiss library~\footnote{https://github.com/facebookresearch/faiss} for index construction. We utilize TransE~\cite{bordes2013translating} to obtain embeddings of entities and relations, and CPT-Text~\cite{neelakantan2022text} to obtain embeddings of questions and memories. The RL agent is implemented with a two-layer neural network, where the hidden layer consists of 20 neurons and uses the tanh activation function. The output layer has 2 neurons corresponding to the action space. Several built-in RL codes are available in~\cite{wang2021trajectory, zhang2023online}. The hyperparameter $K$ for activated nodes is empirically set to 3. We generate 1,000 episodes for the warm-start stage and 100 episodes for the policy gradient stage. We use the Adam stochastic gradient descent with a learning rate of 0.001 to optimize the policy, and the reward discount is set to 0.99. We cache the generated QA pairs~\footnote{https://github.com/zilliztech/GPTCache} during training to boost efficiency.
\subsection{Experimental Results}
\label{sec:result}

\noindent \textbf{(1) Effectiveness evaluation (question answering).} We compare the \texttt{EMG-RAG} with other RAG methods for question answering on three LLMs. As shown in Table~\ref{tab:effectiveness}, we observe that the performance of \texttt{EMG-RAG} consistently outperforms the baselines. For example, it improves upon the best baseline method, M-RAG, by 5.3\%, 8.3\%, 3.9\%, and 18.4\% in terms of R-1, R-2, R-L, and BLEU, respectively. This improvement is due to two main factors: 1) it captures complex relationships between memories with the EMG, and 2) it effectively selects essential memories for the RAG execution. Additionally, GPT-4 demonstrates superior performance compared to other LLMs, and \texttt{EMG-RAG} shows comparable performance to M-RAG even when deployed on the relatively smaller ChatGLM3-6B.

\noindent \textbf{(2) Effectiveness evaluation (autofill forms).} We further evaluate the \texttt{EMG-RAG} for autofill forms, and it shows consistent improvement, as detailed in Table~\ref{tab:effectiveness}. For example, it surpasses M-RAG by 2.2\% in terms of exact match accuracy.

\noindent \textbf{(3) Effectiveness evaluation (user services).} We target two specific domains of user services: 1) reminders of important events and their times, and 2) travel services involving destination addresses for navigation. We report the exact match accuracy for events and times (reminders), and addresses (travel) in Table~\ref{tab:effectiveness}. The improvements over M-RAG for the two tasks are 2.9\% and 5.5\%. 

\noindent \textbf{(4) Effectiveness evaluation (continuous edits).} We evaluate the effectiveness of \texttt{EMG-RAG} in supporting continuous edits over a period of 4 weeks. The results, in terms of R-L for question answering (QA), and exact match accuracy for autofill forms (AF) and user services (US, combining reminder and travel results), are presented in Table~\ref{tab:duration}. We observe that \texttt{EMG-RAG} consistently outperforms M-RAG, by approximately 10.6\%, 9.5\%, and 9.7\% for QA, AF, and US, respectively. This is owing to the editability of \texttt{EMG-RAG}, whereas M-RAG simply incorporates edits into a database, where many memories may become outdated for answering. Additionally, we report the total number of edits involved in the testing set for each week.

\noindent \textbf{(5) Ablation study.} To evaluate the effectiveness of different components in \texttt{EMG-RAG}, we conduct an ablation study. (1) We omit the design of activated nodes, and the search starts from the root of EMG. (2) We remove the warm-start stage (WS) and only train the policy in the policy gradient stage (PG). (3) We remove the PG and use the WS only. For (1), it results in a performance drop (e.g., R-1 from 93.46 to 90.96), because many irrelevant memories (as noises) may be retrieved if the search starts from the root. For (2) and (3), we observe that the PG contributes the most to the result (e.g., R-1 from 93.46 to 90.59), because it can explicitly optimize the performance end-to-end, and WS provides a basic memory selection ability for the agent.

\begin{table}[t]
\caption{Impacts of the number of $K$ for activated nodes.}
\setlength{\tabcolsep}{3pt}
\centering
\vspace{-3mm}
\label{tab:k}
\begin{tabular}{l|ccccc}
\hline
$K$             & 1 & 2 & 3 & 4 & 5 \\ \hline
R-L           &84.55 &86.06 &\textbf{88.06} &88.06 &87.19   \\
Inference (s) &1.35  &1.63  &\textbf{2.14}  &2.55  &3.32   \\ \hline
\end{tabular}
\vspace{-3mm}
\end{table}

\noindent \textbf{(6) Parameter study ($K$ for activated nodes).} We vary the value of $K$ from 1 to 5 and report the R-L score for the question answering task, along with the corresponding inference times. As shown in Table~\ref{tab:k}, we observe that $K=3$ provides the best effectiveness while maintaining reasonable inference time. When $K$ is smaller, the limited number of activated nodes for graph traversal restricts the ability to find crucial memories. Conversely, when $K$ is larger, it activates many nodes and returns numerous memories, potentially introducing noise that hinders the LLM generation. As expected, the inference time increases as $K$ increases.

\begin{table}[t]
\caption{Online A/B test.}
\label{tab:abtest}
\setlength{\tabcolsep}{2pt}
\centering
\vspace{-3mm}
\begin{tabular}{l|ccc}
\hline
\multirow{2}{*}{Apps} & \multicolumn{3}{c}{Cold-start}           \\ \cline{2-4} 
                              & A (old \texttt{EMG-RAG}) & B (new \texttt{EMG-RAG}) & Impr \\ \hline
QA            &88.06    &\textbf{91.99}            &4.5\%      \\
AF            &92.86    &\textbf{95.85}            &3.2\%      \\
US            &94.66    &\textbf{97.56}            &3.1\%      \\ \hline
\end{tabular}
\vspace{-4mm}
\end{table}
  
\noindent \textbf{(7) Online A/B test.} We perform an online A/B test over one month to compare the new system with the existing one. During this period, we collect real users' questions and manually written answers to fine-tune the model. The results, presented in Table~\ref{tab:abtest}, show further improvements across all applications. It highlights a cold-start problem caused by distributional shifts between questions generated by GPT-4 and those posed by real users. We use GPT-4-generated questions for model training because they cover diverse scenarios and allow for the automatic collection of required memories, enabling large-scale training. Once the trained model is deployed, we fine-tune it using real user questions and manually written answers through online learning as described in Section~\ref{sec:apps}.

\begin{table}[]
\caption{Data quality evaluation.}
\vspace{-3mm}
\label{tab:quality}
\begin{tabular}{c|c|c|c}
\hline
Data Quality     & QA & AF & US \\ \hline
Human Evaluation & 91.1\%             & 87.5\%         & 97.4\%        \\
GPT-4 Evaluation                                                & 93.3\%             & 98.7\%         & 99.3\%        \\ \hline
\end{tabular}
\vspace{-5mm}
\end{table}

\noindent\textbf{(8) Data quality evaluation.} We evaluate data quality across three data collection steps. For Step-1, we note that OCR is a well-established technology used to extract information from app screenshots in our study. Given that the printed fonts from apps are typically standard, OCR is not expected to face significant challenges. For Step-2 and Step-3, we utilize the powerful GPT-4 model for memory and QA pair collection and assess quality from two perspectives: (1) Qualitatively: We present memory samples from our focus domains as shown in Table~\ref{atab:subclass}, which generally meet the expected precision. (2) Quantitatively: We assess quality using human evaluation and LLM evaluation. The results are reported in Table~\ref{tab:quality}. For human evaluation, we randomly selected 10\% of the user data and asked five participants to annotate the answers (for QA) and entities (for AF and US) based on the collected questions and memories. By comparing the human-annotated answers and entities with those generated by GPT-4, we report a R-L score of 91.1\% for QA and exact match scores of 87.5\% for AF and 97.4\% for US. These results demonstrate the high accuracy of the collected data. For LLM evaluation, we employ a method where GPT-4 self-verifies whether it can generate answers (or entities) that are consistent with those obtained during the data collection, based on the collected questions and required memories. The evaluation reveals the scores of 93.3\%, 98.7\%, and 99.3\% for the three applications, respectively, demonstrating a high level of consistency and effectiveness.


\section{Conclusion}
\label{sec:conclusion}

In this paper, we present a novel task of creating personalized agents powered by LLMs, which leverage users' personal memories to enhance three downstream applications. Our solution, \texttt{EMG-RAG}, combines RAG techniques with an EMG to tackle challenges in data collection, editability, and selectability. Extensive experiments are conducted to confirm the effectiveness of \texttt{EMG-RAG}.

\section{Limitations}
For limitations, while only the parameters of the RL agent are trained and the parameters of the LLMs remain fixed, the training efficiency is not higher than that of a Naive RAG setup. This inefficiency stems from the need to query the LLM during training to obtain answers for optimization.

\bibliography{ref}

\begin{thebibliography}{47}
\expandafter\ifx\csname natexlab\endcsname\relax\def\natexlab#1{#1}\fi

\bibitem[{Bordes et~al.(2013)Bordes, Usunier, Garcia-Duran, Weston, and Yakhnenko}]{bordes2013translating}
Antoine Bordes, Nicolas Usunier, Alberto Garcia-Duran, Jason Weston, and Oksana Yakhnenko. 2013.
\newblock Translating embeddings for modeling multi-relational data.
\newblock \emph{NeurIPS}, 26.

\bibitem[{Briakou et~al.(2023)Briakou, Cherry, and Foster}]{briakou2023searching}
Eleftheria Briakou, Colin Cherry, and George Foster. 2023.
\newblock Searching for needles in a haystack: On the role of incidental bilingualism in palm's translation capability.
\newblock \emph{arXiv preprint arXiv:2305.10266}.

\bibitem[{Cordella et~al.(2004)Cordella, Foggia, Sansone, and Vento}]{cordella2004sub}
Luigi~P Cordella, Pasquale Foggia, Carlo Sansone, and Mario Vento. 2004.
\newblock A (sub) graph isomorphism algorithm for matching large graphs.
\newblock \emph{IEEE TPAMI}, 26(10):1367--1372.

\bibitem[{Das et~al.(2021)Das, Zaheer, Thai, Godbole, Perez, Lee, Tan, Polymenakos, and Mccallum}]{das2021case}
Rajarshi Das, Manzil Zaheer, Dung Thai, Ameya Godbole, Ethan Perez, Jay~Yoon Lee, Lizhen Tan, Lazaros Polymenakos, and Andrew Mccallum. 2021.
\newblock Case-based reasoning for natural language queries over knowledge bases.
\newblock In \emph{EMNLP}, pages 9594--9611.

\bibitem[{De~Cao et~al.(2021)De~Cao, Aziz, and Titov}]{de2021editing}
Nicola De~Cao, Wilker Aziz, and Ivan Titov. 2021.
\newblock Editing factual knowledge in language models.
\newblock In \emph{EMNLP}, pages 6491--6506.

\bibitem[{Du et~al.(2022)Du, Qian, Liu, Ding, Qiu, Yang, and Tang}]{du2022glm}
Zhengxiao Du, Yujie Qian, Xiao Liu, Ming Ding, Jiezhong Qiu, Zhilin Yang, and Jie Tang. 2022.
\newblock Glm: General language model pretraining with autoregressive blank infilling.
\newblock In \emph{ACL}, pages 320--335.

\bibitem[{Han et~al.(2024)Han, McInroe, Jelley, Albrecht, Bell, and Storkey}]{han2024llm}
Dongge Han, Trevor McInroe, Adam Jelley, Stefano~V Albrecht, Peter Bell, and Amos Storkey. 2024.
\newblock Llm-personalize: Aligning llm planners with human preferences via reinforced self-training for housekeeping robots.
\newblock \emph{arXiv preprint arXiv:2404.14285}.

\bibitem[{Hongru et~al.(2023)Hongru, Wang, Mi, Deng, Zezhong, Liang, Xu, and Wong}]{hongru2023cue}
WANG Hongru, Rui Wang, Fei Mi, Yang Deng, WANG Zezhong, Bin Liang, Ruifeng Xu, and Kam-Fai Wong. 2023.
\newblock Cue-cot: Chain-of-thought prompting for responding to in-depth dialogue questions with llms.
\newblock In \emph{EMNLP (Findings)}, pages 12047--12064.

\bibitem[{Hu et~al.(2023)Hu, Iscen, Sun, Wang, Chang, Sun, Schmid, Ross, and Fathi}]{hu2023reveal}
Ziniu Hu, Ahmet Iscen, Chen Sun, Zirui Wang, Kai-Wei Chang, Yizhou Sun, Cordelia Schmid, David~A Ross, and Alireza Fathi. 2023.
\newblock Reveal: Retrieval-augmented visual-language pre-training with multi-source multimodal knowledge memory.
\newblock In \emph{CVPR}, pages 23369--23379.

\bibitem[{Jiang et~al.(2023)Jiang, Zhou, Dong, Ye, Zhao, and Wen}]{jiang2023structgpt}
Jinhao Jiang, Kun Zhou, Zican Dong, Keming Ye, Wayne~Xin Zhao, and Ji-Rong Wen. 2023.
\newblock Structgpt: A general framework for large language model to reason over structured data.
\newblock \emph{arXiv preprint arXiv:2305.09645}.

\bibitem[{Kim et~al.(2020)Kim, Kim, and Kim}]{kim2020will}
Hyunwoo Kim, Byeongchang Kim, and Gunhee Kim. 2020.
\newblock Will i sound like me? improving persona consistency in dialogues through pragmatic self-consciousness.
\newblock In \emph{EMNLP}, pages 904--916.

\bibitem[{Lewis et~al.(2020)Lewis, Perez, Piktus, Petroni, Karpukhin, Goyal, K{\"u}ttler, Lewis, Yih, Rockt{\"a}schel et~al.}]{lewis2020retrieval}
Patrick Lewis, Ethan Perez, Aleksandra Piktus, Fabio Petroni, Vladimir Karpukhin, Naman Goyal, Heinrich K{\"u}ttler, Mike Lewis, Wen-tau Yih, Tim Rockt{\"a}schel, et~al. 2020.
\newblock Retrieval-augmented generation for knowledge-intensive nlp tasks.
\newblock \emph{NeurIPS}, 33:9459--9474.

\bibitem[{Li et~al.(2024{\natexlab{a}})Li, Li, Song, Yang, Ma, and Yu}]{li2024pmet}
Xiaopeng Li, Shasha Li, Shezheng Song, Jing Yang, Jun Ma, and Jie Yu. 2024{\natexlab{a}}.
\newblock Pmet: Precise model editing in a transformer.
\newblock In \emph{AAAI}, volume~38, pages 18564--18572.

\bibitem[{Li et~al.(2024{\natexlab{b}})Li, Wen, Wang, Li, Yuan, Liu, Liu, Xu, Wang, Sun et~al.}]{li2024personal}
Yuanchun Li, Hao Wen, Weijun Wang, Xiangyu Li, Yizhen Yuan, Guohong Liu, Jiacheng Liu, Wenxing Xu, Xiang Wang, Yi~Sun, et~al. 2024{\natexlab{b}}.
\newblock Personal llm agents: Insights and survey about the capability, efficiency and security.
\newblock \emph{arXiv preprint arXiv:2401.05459}.

\bibitem[{Lin(2004)}]{lin-2004-rouge}
Chin-Yew Lin. 2004.
\newblock {ROUGE}: A package for automatic evaluation of summaries.
\newblock In \emph{Text Summarization Branches Out}, pages 74--81.

\bibitem[{Liu et~al.(2020)Liu, Chen, Chen, Lou, Chen, Zhou, and Zhang}]{liu2020you}
Qian Liu, Yihong Chen, Bei Chen, Jian-Guang Lou, Zixuan Chen, Bin Zhou, and Dongmei Zhang. 2020.
\newblock You impress me: Dialogue generation via mutual persona perception.
\newblock In \emph{ACL}, pages 1417--1427.

\bibitem[{Ma et~al.(2023)Ma, Gong, He, Zhao, and Duan}]{DBLP:journals/corr/abs-2305-14283}
Xinbei Ma, Yeyun Gong, Pengcheng He, Hai Zhao, and Nan Duan. 2023.
\newblock Query rewriting for retrieval-augmented large language models.
\newblock \emph{EMNLP}, pages 5303--5315.

\bibitem[{Meng et~al.(2022{\natexlab{a}})Meng, Bau, Andonian, and Belinkov}]{meng2022locating}
Kevin Meng, David Bau, Alex Andonian, and Yonatan Belinkov. 2022{\natexlab{a}}.
\newblock Locating and editing factual associations in gpt.
\newblock \emph{NeurIPS}, 35:17359--17372.

\bibitem[{Meng et~al.(2022{\natexlab{b}})Meng, Sharma, Andonian, Belinkov, and Bau}]{meng2022mass}
Kevin Meng, Arnab~Sen Sharma, Alex Andonian, Yonatan Belinkov, and David Bau. 2022{\natexlab{b}}.
\newblock Mass-editing memory in a transformer.
\newblock \emph{arXiv preprint arXiv:2210.07229}.

\bibitem[{Mitchell et~al.(2021)Mitchell, Lin, Bosselut, Finn, and Manning}]{mitchell2021fast}
Eric Mitchell, Charles Lin, Antoine Bosselut, Chelsea Finn, and Christopher~D Manning. 2021.
\newblock Fast model editing at scale.
\newblock \emph{arXiv preprint arXiv:2110.11309}.

\bibitem[{Mitchell et~al.(2022)Mitchell, Lin, Bosselut, Manning, and Finn}]{mitchell2022memory}
Eric Mitchell, Charles Lin, Antoine Bosselut, Christopher~D Manning, and Chelsea Finn. 2022.
\newblock Memory-based model editing at scale.
\newblock In \emph{ICML}, pages 15817--15831. PMLR.

\bibitem[{Neelakantan et~al.(2022)Neelakantan, Xu, Puri, Radford, Han, Tworek, Yuan, Tezak, Kim, Hallacy et~al.}]{neelakantan2022text}
Arvind Neelakantan, Tao Xu, Raul Puri, Alec Radford, Jesse~Michael Han, Jerry Tworek, Qiming Yuan, Nikolas Tezak, Jong~Wook Kim, Chris Hallacy, et~al. 2022.
\newblock Text and code embeddings by contrastive pre-training.
\newblock \emph{CoRR}.

\bibitem[{OpenAI(2023)}]{OpenAI2023GPT-4}
OpenAI. 2023.
\newblock {GPT-4} technical report.
\newblock \emph{arXiv preprint}.

\bibitem[{Post(2018)}]{DBLP:conf/wmt/Post18}
Matt Post. 2018.
\newblock A call for clarity in reporting {BLEU} scores.
\newblock In \emph{{WMT}}, pages 186--191.

\bibitem[{Radford et~al.(2018)Radford, Narasimhan, Salimans, Sutskever et~al.}]{radford2018improving}
Alec Radford, Karthik Narasimhan, Tim Salimans, Ilya Sutskever, et~al. 2018.
\newblock Improving language understanding by generative pre-training.

\bibitem[{Ren et~al.(2023)Ren, Zhou, Meng, Huang, Wang, Wang, Li, Zhang, Podolskiy, Arshinov et~al.}]{ren2023pangu}
Xiaozhe Ren, Pingyi Zhou, Xinfan Meng, Xinjing Huang, Yadao Wang, Weichao Wang, Pengfei Li, Xiaoda Zhang, Alexander Podolskiy, Grigory Arshinov, et~al. 2023.
\newblock Pangu-$\{$$\backslash$Sigma$\}$: Towards trillion parameter language model with sparse heterogeneous computing.
\newblock \emph{arXiv preprint arXiv:2303.10845}.

\bibitem[{Ribeiro et~al.(2021)Ribeiro, Paredes, Silva, Aparicio, and Silva}]{ribeiro2021survey}
Pedro Ribeiro, Pedro Paredes, Miguel~EP Silva, David Aparicio, and Fernando Silva. 2021.
\newblock A survey on subgraph counting: concepts, algorithms, and applications to network motifs and graphlets.
\newblock \emph{ACM Computing Surveys (CSUR)}, 54(2):1--36.

\bibitem[{Shu et~al.(2022)Shu, Yu, Li, Karlsson, Ma, Qu, and Lin}]{shu2022tiara}
Yiheng Shu, Zhiwei Yu, Yuhan Li, B{\"o}rje Karlsson, Tingting Ma, Yuzhong Qu, and Chin-Yew Lin. 2022.
\newblock Tiara: Multi-grained retrieval for robust question answering over large knowledge base.
\newblock In \emph{EMNLP}, pages 8108--8121.

\bibitem[{Silver et~al.(2014)Silver, Lever, Heess, Degris, Wierstra, and Riedmiller}]{silver2014deterministic}
David Silver, Guy Lever, Nicolas Heess, Thomas Degris, Daan Wierstra, and Martin Riedmiller. 2014.
\newblock Deterministic policy gradient algorithms.
\newblock In \emph{ICML}, pages 387--395. PMLR.

\bibitem[{Song et~al.(2021)Song, Wang, Zhang, Zhang, and Liu}]{song2021bob}
Haoyu Song, Yan Wang, Kaiyan Zhang, Weinan Zhang, and Ting Liu. 2021.
\newblock Bob: Bert over bert for training persona-based dialogue models from limited personalized data.
\newblock In \emph{ACL}, pages 167--177.

\bibitem[{Ullmann(1976)}]{ullmann1976algorithm}
Julian~R Ullmann. 1976.
\newblock An algorithm for subgraph isomorphism.
\newblock \emph{Journal of the ACM (JACM)}, 23(1):31--42.

\bibitem[{Wang et~al.(2023{\natexlab{a}})Wang, Xu, Peng, Zhang, Chen, Wang, Feng, and An}]{wang2023keqing}
Chaojie Wang, Yishi Xu, Zhong Peng, Chenxi Zhang, Bo~Chen, Xinrun Wang, Lei Feng, and Bo~An. 2023{\natexlab{a}}.
\newblock keqing: knowledge-based question answering is a nature chain-of-thought mentor of llm.
\newblock \emph{arXiv preprint arXiv:2401.00426}.

\bibitem[{Wang et~al.(2023{\natexlab{b}})Wang, Hu, Deng, Wang, Mi, Wang, Wang, Kwan, King, and Wong}]{wang2023large}
Hongru Wang, Minda Hu, Yang Deng, Rui Wang, Fei Mi, Weichao Wang, Yasheng Wang, Wai~Chung Kwan, Irwin King, and Kam-Fai Wong. 2023{\natexlab{b}}.
\newblock Large language models as source planner for personalized knowledge-grounded dialogues.
\newblock In \emph{EMNLP (Findings)}, pages 9556--9569.

\bibitem[{Wang et~al.(2021)Wang, Long, and Cong}]{wang2021trajectory}
Zheng Wang, Cheng Long, and Gao Cong. 2021.
\newblock Trajectory simplification with reinforcement learning.
\newblock In \emph{ICDE}, pages 684--695. IEEE.

\bibitem[{Wang et~al.(2024)Wang, Teo, Ouyang, Xu, and Shi}]{wang2024mrag}
Zheng Wang, Shu~Xian Teo, Jieer Ouyang, Yongjun Xu, and Wei Shi. 2024.
\newblock M-{RAG}: Reinforcing large language model performance through retrieval-augmented generation with multiple partitions.
\newblock In \emph{ACL}.

\bibitem[{Welch et~al.(2022)Welch, Gu, Kummerfeld, P{\'e}rez-Rosas, and Mihalcea}]{welch2022leveraging}
Charles Welch, Chenxi Gu, Jonathan~K Kummerfeld, Ver{\'o}nica P{\'e}rez-Rosas, and Rada Mihalcea. 2022.
\newblock Leveraging similar users for personalized language modeling with limited data.
\newblock In \emph{ACL}, pages 1742--1752.

\bibitem[{Williams(1992)}]{williams1992simple}
Ronald~J Williams. 1992.
\newblock Simple statistical gradient-following algorithms for connectionist reinforcement learning.
\newblock \emph{Machine learning}, 8(3):229--256.

\bibitem[{Xu et~al.(2022{\natexlab{a}})Xu, Li, Wang, Yang, Wang, and Xiao}]{xu2022cosplay}
Chen Xu, Piji Li, Wei Wang, Haoran Yang, Siyun Wang, and Chuangbai Xiao. 2022{\natexlab{a}}.
\newblock Cosplay: Concept set guided personalized dialogue generation across both party personas.
\newblock In \emph{SIGIR}, pages 201--211.

\bibitem[{Xu et~al.(2022{\natexlab{b}})Xu, Gou, Wu, Niu, Wu, Wang, and Wang}]{xu2022long}
Xinchao Xu, Zhibin Gou, Wenquan Wu, Zheng-Yu Niu, Hua Wu, Haifeng Wang, and Shihang Wang. 2022{\natexlab{b}}.
\newblock Long time no see! open-domain conversation with long-term persona memory.
\newblock In \emph{ACL (Findings)}, pages 2639--2650.

\bibitem[{Yang et~al.(2023)Yang, Chen, Wang, Hu, and Zhang}]{yang2023enhancing}
Qian Yang, Qian Chen, Wen Wang, Baotian Hu, and Min Zhang. 2023.
\newblock Enhancing multi-modal multi-hop question answering via structured knowledge and unified retrieval-generation.
\newblock In \emph{ACM MM}, pages 5223--5234.

\bibitem[{Ye et~al.(2021)Ye, Yavuz, Hashimoto, Zhou, and Xiong}]{ye2021rng}
Xi~Ye, Semih Yavuz, Kazuma Hashimoto, Yingbo Zhou, and Caiming Xiong. 2021.
\newblock Rng-kbqa: Generation augmented iterative ranking for knowledge base question answering.
\newblock \emph{arXiv preprint arXiv:2109.08678}.

\bibitem[{Zhang et~al.(2020)Zhang, Liu, Xiong, and Liu}]{zhang2020grounded}
Houyu Zhang, Zhenghao Liu, Chenyan Xiong, and Zhiyuan Liu. 2020.
\newblock Grounded conversation generation as guided traverses in commonsense knowledge graphs.
\newblock In \emph{ACL}, pages 2031--2043.

\bibitem[{Zhang et~al.(2023{\natexlab{a}})Zhang, Zhao, Kang, and Liu}]{zhang2023memory}
Kai Zhang, Fubang Zhao, Yangyang Kang, and Xiaozhong Liu. 2023{\natexlab{a}}.
\newblock Memory-augmented llm personalization with short-and long-term memory coordination.
\newblock \emph{arXiv preprint arXiv:2309.11696}.

\bibitem[{Zhang et~al.(2023{\natexlab{b}})Zhang, Wang, Long, Huang, Yiu, Liu, Cong, and Shi}]{zhang2023online}
Qianru Zhang, Zheng Wang, Cheng Long, Chao Huang, Siu-Ming Yiu, Yiding Liu, Gao Cong, and Jieming Shi. 2023{\natexlab{b}}.
\newblock Online anomalous subtrajectory detection on road networks with deep reinforcement learning.
\newblock In \emph{ICDE}, pages 246--258. IEEE.

\bibitem[{Zhang et~al.(2018)Zhang, Dinan, Urbanek, Szlam, Kiela, and Weston}]{zhang2018personalizing}
Saizheng Zhang, Emily Dinan, Jack Urbanek, Arthur Szlam, Douwe Kiela, and Jason Weston. 2018.
\newblock Personalizing dialogue agents: I have a dog, do you have pets too?
\newblock In \emph{ACL}, pages 2204--2213.

\bibitem[{Zhao et~al.(2024)Zhao, Zhang, Yu, Wang, Geng, Fu, Yang, Zhang, and Cui}]{zhao2024retrieval}
Penghao Zhao, Hailin Zhang, Qinhan Yu, Zhengren Wang, Yunteng Geng, Fangcheng Fu, Ling Yang, Wentao Zhang, and Bin Cui. 2024.
\newblock Retrieval-augmented generation for ai-generated content: A survey.
\newblock \emph{arXiv preprint arXiv:2402.19473}.

\bibitem[{Zhong et~al.(2024)Zhong, Guo, Gao, Ye, and Wang}]{zhong2024memorybank}
Wanjun Zhong, Lianghong Guo, Qiqi Gao, He~Ye, and Yanlin Wang. 2024.
\newblock Memorybank: Enhancing large language models with long-term memory.
\newblock In \emph{AAAI}, volume~38, pages 19724--19731.

\end{thebibliography}
\bibliographystyle{acl_natbib}

\clearpage
\appendix \section{Appendix}
\label{sec:appendix}

\subsection{Data Collection Details}
\label{asec:data_collection}
The data collection process involves three key steps, which are presented below:

\textbf{Step-1: Raw Data Collection.} We explore two approaches, termed Active Remember (AR) and Passive Remember (PR), for collecting raw data derived from users' daily conversations with AI assistants and screenshots from their apps. With AR, the AI assistant is trained to actively classify data (such as conversation sentences) into supported subclasses outlined in Table~\ref{atab:subclass}, and filter out noise data. With PR, users have the option to directly let the assistant to remember specific content for future use. Leveraging AR and PR, we remove a significant volume of trivial data, and then extract memories from the refined dataset. 

\textbf{Step-2: Memory Data Construction.} We utilize a LLM, such as GPT-4, with the refined dataset to generate structured memories from the raw data. Additionally, we integrate various natural language processing techniques, including absolute date and time conversion, entity anaphora resolution, and event coreference resolution, to further clean the memories and facilitate graph construction.

\textbf{Step-3: QA Pairs Construction.} We organize the memory data chronologically and partition it into separate conversation sessions. Then, a LLM generates QA pairs for each session. To create complex questions for targeted training, such as those requiring multiple memories for answering, we explicitly instruct the LLM to utilize multiple associative relationships between memories to generate questions, ensuring that at least one or more memories are needed for accurate responses.

\subsection{Memory Types and Subclasses}
\label{asec:MSL}
We describe the 4 memory types: (1) Relationship, which involves recognizing users' surrounding relationships and attributes of related individuals, such as birthdays and names of family members; (2) Preference, where we identify users' likes and dislikes for various topics or entities; (3) Event, focusing on key event information about users, such as their status, recent experiences, and upcoming schedules; and (4) Attribute, encompassing users' personal details such as name, gender, age, possessions, and other relevant information. 

We enumerate the supported business subclasses of the EMG with memory examples in Table~\ref{atab:subclass}.

\begin{table*}[]
\small
\setlength{\tabcolsep}{6pt}
\caption{The supported memory subclasses with memory examples.}
\label{atab:subclass}
\begin{tabular}{|l|l|l|}
\hline
Memory Types                  & Memory Subclasses                                                                                                          & Memory Examples                                                                                                                                                      \\ \hline
\multirow{5}{*}{Relationship} & Spouse                                                                                                                     & \multirow{5}{*}{Tomorrow is my mom's birthday.}                                                                                                               \\ \cline{2-2}
                              & Parents/Children                                                                                                           &                                                                                                                                                               \\ \cline{2-2}
                              & Relatives                                                                                                                  &                                                                                                                                                               \\ \cline{2-2}
                              & Colleague/Friends                                                                                                          &                                                                                                                                                               \\ \cline{2-2}
                              & Teacher/Student                                                                                                            &                                                                                                                                                               \\ \hline
\multirow{6}{*}{Preference}   & Diet preference                                                                                                            & I like spicy food.                                                                                                                                            \\ \cline{2-3} 
                              & Cultural preference (tourism, travel)                                                                                      & \begin{tabular}[c]{@{}l@{}}I enjoy traveling by airplane.\\ I like going to museums.\end{tabular}                                                             \\ \cline{2-3} 
                              & Car preference                                                                                                             & I like BMWs.                                                                                                                                                  \\ \cline{2-3} 
                              & \begin{tabular}[c]{@{}l@{}}Sports preference\\ (favorite sports types, sports celebrities)\end{tabular}                    & \begin{tabular}[c]{@{}l@{}}I like playing table tennis on weekends.\\ James is my favorite basketball star.\end{tabular}                                      \\ \cline{2-3} 
                              & Gaming preference (category, name)                                                                                         & I like the game League of Legends.                                                                                                                            \\ \cline{2-3} 
                              & \begin{tabular}[c]{@{}l@{}}Audio-visual entertainment preference\\ (favorite videos, music, movies, TV shows)\end{tabular} & \begin{tabular}[c]{@{}l@{}}I like science fiction movies.\\ I like listening to Jay Chou's songs.\end{tabular}                                                \\ \hline
\multirow{3}{*}{Event}        & \begin{tabular}[c]{@{}l@{}}Life events\\ (academic, marriage, buying a flat, parenting)\end{tabular}                       & \begin{tabular}[c]{@{}l@{}}The college entrance examination is coming soon.\\ I met a girlfriend online.\\ My family is welcoming a second child.\end{tabular} \\ \cline{2-3} 
                              & Arrangement                                                                                                                & \begin{tabular}[c]{@{}l@{}}I'm going to visit clients tomorrow.\\ I want to travel to Amsterdam next month.\\ I have an oral defense next Monday.\end{tabular}      \\ \cline{2-3} 
                              & Anniversary                                                                                                                & Next month's fifth is our wedding anniversary.                                                                                                                \\ \hline
\multirow{7}{*}{Attribute}    & Name/Nickname                                                                                                              & My name is Wang Xiaoming, call me Lord Radish.                                                                                                                \\ \cline{2-3} 
                              & Birthday/Age                                                                                                               & \begin{tabular}[c]{@{}l@{}}I am 17 years old this year.\\ I was born in 1998.\\ My birthday is April 2nd.\end{tabular}                                        \\ \cline{2-3} 
                              & Gender                                                                                                                     & I am a girl.                                                                                                                                                  \\ \cline{2-3} 
                              & Education                                                                                                                  & I am an undergraduate student.                                                                                                                                \\ \cline{2-3} 
                              & Personal belongings/Pets                                                                                                   & Riding my beloved electric scooter, my pink BMW.                                                                                                              \\ \cline{2-3} 
                              & Address                                                                                                                    & I reside in Jurong West, Singapore.                                                                                                                                                \\ \cline{2-3} 
                              & Occupation                                                                                                                 & I am a research scientist.                                                                                                                                    \\ \hline
\end{tabular}
\end{table*}

\subsection{Further Discussion}
\label{asec:discuss} 

\noindent\textbf{Q1. The necessity of using a graph if the user memory size is small.}

We analyze the user memory size based on statistical data. We list the number of memories generated from user interactions with the intelligent assistant over the past one day, in descending order: the Top-1,000 user has 296 memories; the Top-10,000 user has 101 memories; and the Top-20,000 user has 72 memories. Notably, around 20,000 users produce at least 70 memories each day, and the memory size increases over time. These users represent a significant portion that should not be overlooked, necessitating a graph structure (such as EMG) for effective management.

Moreover, using a graph enhances effectiveness by naturally capturing semantic relationships between memories, which improves reasoning during RAG. As demonstrated by the experimental results in Table~\ref{tab:effectiveness} and Table~\ref{tab:duration}, our approach outperforms the NiaH, Naive, and M-RAG baselines, achieving approximately a 10\% improvement over the best baseline M-RAG, which manages the memory instances independently.

\noindent\textbf{Q2. What would be the storage and computation costs of EMG if the number of users is larger? Is it possible to share some common patterns of different users in this design?}

We clarify that EMGs are independently managed and computed on each user's personal device, and the storage and computation costs are not impacted by the number of users in practice.
To reduce storage costs, we consider a potential solution of sharing common patterns across different users' EMGs. We aim to mine common subgraph patterns using classic subgraph isomorphism algorithms~\cite{ribeiro2021survey,ullmann1976algorithm,cordella2004sub}. On the server side, we will manage the common patterns identified by Graph ID (GID), and link user-specific data identified by User ID (UID) to them. On the user side, GIDs and UIDs will replace the corresponding data in the EMGs, minimizing duplication across different users' EMGs.

\noindent\textbf{Q3. Privacy discussions in data collection and model training, e.g., is the model trained in a single user-based? Will the model output other user's information during inference?}

In data collection, we clarify that the data is collected solely for individual use to provide relevant applications. Each user’s EMG is independently built based on the collected data, and will not be shared. Additionally, all user information presented in this research has been de-identified.

In model training, we use a single-user approach and address privacy concerns as follows: (1) EMGs are independently managed for each user and are not shared. (2) The RL agent is a simple neural network that includes or excludes nodes (actions 1 or 0) in the personal graph. (3) The LLM remains frozen, ensuring it does not memorize user data or output information from other users.

\subsection{Prompts for Data Collection}
\label{asec:prompts}
Table~\ref{tab:raw_to_mem} presents the prompt for collecting memories from raw extracted data, while Table~\ref{tab:reasoning} provides the prompt for generating reasoning as the required memories for QA pairs. The prompt for generating QA pairs based on this reasoning is presented in Table~\ref{tab:qa}. Additionally, Table~\ref{tab:mem_syn} offers an alternative method to synthesize memories when raw extracted data is unavailable.

\begin{table*}[]
\caption{Prompt for collecting memories from raw extracted data.}
\label{tab:raw_to_mem}
\begin{tabular}{p{15cm}}
\hline
Please help me organize the following raw user data into standardized memory data.\\\\

Here is an example format:\\
\quad 1. My name is Zhang Zhenqiang.\\
\quad 2. My zodiac sign is Aquarius.\\
\quad 3. My company's address is Oriental International, Pudong New District, Shanghai.\\
\quad 4. My mother's birthday is April 8, 1982.\\
\quad 5. My father's favorite sport is basketball.\\
\quad 6. I watched the movie ``Fast and Furious'' at Orange Cinema in July 2023.\\
\quad 7. Next Saturday, I will attend a high school friend's wedding.\\\\

Note: Please use the above format to output and display all the data. + \{raw data\}
\\ \hline
\end{tabular}
\end{table*}

\begin{table*}[]
\caption{Prompt for generating reasoning as the required memories for QA pairs.}
\label{tab:reasoning}
\begin{tabular}{p{15cm}}
\hline
You have many memories from one person. Explore all possible associations, including multi-hop and connections around the same event, person, or entity. Records can intersect between different associations.\\\\

Here is an example: Assume the following memory records exist:\\

\{``ID'': 1, ``Memory Content'': ``Recently, my sleep hasn't been good and lacks deep sleep.'', ``Memory Location'': ``Lychee Garden Apartment, Longgang District, Shenzhen, Guangdong Province'', ``Memory Time'': ``2024-04-22 08:31:19''\}\\

\{``ID'': 2, ``Memory Content'': ``My girlfriend likes to eat durian.'', ``Memory Location'': ``Wuhe Avenue, Longgang District, Shenzhen, Guangdong Province'', ``Memory Time'': ``2021-11-14 15:31:54''\}\\

\{``ID'': 3, ``Memory Content'': ``Baiguoyuan is having a durian promotion next week, and I want to buy some.'', ``Memory Location'': ``Tianan Cloud Valley Building 1, B Section, Xuegang North Road, Longgang District, Shenzhen, Guangdong Province'', ``Memory Time'': ``2022-10-10 09:30:27''\}\\\\

Based on the above memory records, you can extract multiple sets of associations as follows:\\

The memory mentions that your girlfriend likes to eat durian (Memory Point 2), and Baiguoyuan is having a durian promotion next week (Memory Point 3). You could buy some durian from Baiguoyuan during the promotion, so these memories are related (Memory Points 2|3).\\\\

Please extract the associations from the following memory records as thoroughly as possible based on the above example. Multi-hop reasoning relationships, associations around the same event, person, or entity can all be considered as existing connections. Memory records within each association group can intersect; for example, a memory point appearing in one set of associations can also appear in another set if it is reasonable. Please meet the above requirements and return the output following the example. + \{memories\}
\\ \hline
\end{tabular}
\end{table*}

\begin{table*}[]
\caption{Prompt for generating QA pairs.}
\label{tab:qa}
\begin{tabular}{p{15cm}}
\hline
You currently have a set of historical memory records from the same mobile user and hints of multiple associations between these memories. Based on all the memory information and their associations, design some intent statements or questions with the corresponding answers outputting as <questions, answers> for the mobile assistant that require at least one memory record to provide an accurate response. Below is an example:\\\\

Example:

Given the following memory records:

\{``ID'': 1, ``Memory Content'': ``Recently, my sleep hasn't been good and lacks deep sleep.'', ``Memory Location'': ``Lychee Garden Apartment, Longgang District, Shenzhen, Guangdong Province'', ``Memory Time'': ``2024-04-22 08:31:19''\}

\{``ID'': 2, ``Memory Content'': ``My girlfriend likes to eat durian.'', ``Memory Location'': ``Wuhe Avenue, Longgang District, Shenzhen, Guangdong Province'', ``Memory Time'': ``2021-11-14 15:31:54''\}

\{``ID'': 3, ``Memory Content'': ``Baiguoyuan is having a durian promotion next week, and I want to buy some.'', ``Memory Location'': ``Tianan Cloud Valley Building 1, B Section, Xuegang North Road, Longgang District, Shenzhen, Guangdong Province'', ``Memory Time'': ``2022-10-10 09:30:27''\}\\\\

Based on the above memory information, there are the following association hints:

Your girlfriend likes durian (Memory 2), and Baiguoyuan has a durian promotion next week (Memory 3). You could buy some durian at Baiguoyuan during the promotion. These memories are related around durian (Memory 2|3).\\\\

Based on the above memory information and associations, you can construct the following intent statements or questions:

``I want to buy something delicious for my girlfriend. Any recommendations?'' (Requires Memory 2|3)

The corresponding answers:

``You could buy some durian at Baiguoyuan during the promotion''\\\\

Please construct intent statements or questions from the following memory information and associations, meeting all of the following requirements:

\quad 1. The statements or questions should be directed from the user to the mobile assistant, not questions from the assistant to the user (important requirement).

\quad 2. They should require at least one memory record to provide an accurate response (important requirement).

\quad 3. Keep the content concise and avoid including details already mentioned in the memory records (important requirement).

\quad 4. Avoid intent statements or questions related to reminders (important requirement).

\quad 5. Include both questions and casual statements (important requirement).

\quad 6. End with the required memory points for response in parentheses (important requirement).\\\\

The memory information is as follows: \{memories\}

The memory association hints are as follows: \{reasoning\}
\\ \hline
\end{tabular}
\end{table*}

\begin{table*}[]
\caption{Prompt for synthesizing memories.}
\label{tab:mem_syn}
\begin{tabular}{p{15cm}}
\hline
Please act as a conversation context manager and help me generate personal memory-related data. Below are some examples; please use them as a reference for generating memory data:\\

\{``ID'': 1, ``Memory Content'': ``My girlfriend likes to eat durian.'', ``Memory Location'': ``Wuhe Avenue, Longgang District, Shenzhen, Guangdong Province'', ``Memory Time'': ``2021-11-14 15:31:54''\}

\{``ID'': 2, ``Memory Content'': ``Baiguoyuan is having a durian promotion next week, and I want to buy some.'', ``Memory Location'': ``Tianan Cloud Valley Building 1, B Section, Xuegang North Road, Longgang District, Shenzhen, Guangdong Province'', ``Memory Time'': ``2022-10-10 09:30:27''\}\\\\\

The generated data should meet the following requirements:

\quad 1. The memory data is generated from conversations with a mobile assistant and reflects everyday scenarios. The protagonist is a single individual. You may create realistic content including, but not limited to, basic information about the individual and their family and friends (birthdays, anniversaries, zodiac signs, ID information, passport information, bank card information, etc.), events (meetings, gatherings, travels, renovations, etc.), and order information (movie tickets, hotel reservations, train tickets, flight tickets, etc.). Make sure there are no logical conflicts between the generated data.\\

\quad 2. The memories should exhibit logical multi-step reasoning and not be completely unrelated. For example: “My mom's older brother is named Li Aiguo” and “My uncle's address is a small shop next to Tiananmen Square.” These two memories are linked through the fact that my mom's older brother (my uncle) acts as a reasoning hub, allowing me to deduce that Li Aiguo's address is the small shop next to Tiananmen Square.\\

\quad 3. Ensure that there are no real-world logical conflicts between memory content, locations, and times. For example, earlier memories should have earlier timestamps than later ones. Avoid generating two memories with locations far apart within a short timeframe, such as a memory from Beijing at 20:15:47 and another from Guangzhou at 20:16:20 on the same day.\\

\quad 4. Memory locations can include scenarios like business trips and travel; they do not all need to be in the same city. Memories can be generated in multiple cities.\\\\

Generate 50 more memories in JSONL format, numbered 1 to 50, with each entry including ``memory content'', ``memory location'', and ``memory time''.
\\ \hline
\end{tabular}
\end{table*}

\end{document}